\documentclass[runningheads]{llncs}
\usepackage[T1]{fontenc}

\usepackage{graphicx,verbatim,amssymb,amsmath,algorithm}
\usepackage{algorithmic}
\usepackage{arydshln} 
\usepackage{colortbl}
\usepackage{bbding}
\usepackage{xcolor,xspace}

\usepackage{hyperref}
\usepackage{color}

\def\name{BeerLaNet\xspace}

\def\X{\mathbf{X}}
\def\S{\mathbf{S}}
\def\D{\mathbf{D}}
\def\s{\mathbf{s}}
\def\d{\mathbf{d}}
\def\x{\mathbf{x}}
\def\1{\mathbf{1}}

\def\Re{\mathbb{R}}

\newcommand{\myparagraph}[1]{\smallskip \noindent \textbf{#1}}

\begin{document}

\title{Adaptive Stain Normalization for Cross-Domain Medical Histology}
%
% ==========

\author{
    Tianyue Xu\inst{1}\textsuperscript{$\star$} %index{Xu, Tianyue}
    \and
    Yanlin Wu\inst{1}\textsuperscript{$\star$} %index{Wu, Yanlin}
    \and
    Abhai K. Tripathi\inst{2} %index{Tripathi, Abhai}
    \and
    Matthew M. Ippolito\inst{2,3}\textsuperscript{$\dagger$} %index{Ippolito, Matthew}
    \and
    Benjamin D. Haeffele\inst{4}\textsuperscript{$\dagger$}\Envelope %index{Haeffele, Benjamin}
}

\makeatletter
\renewcommand\@makefnmark{}
\makeatother

\footnotetext[0]{\textsuperscript{$\star$} Equal contribution, co-first authors.}
\footnotetext[0]{\textsuperscript{$\dagger$} Equal contribution, co-senior authors.}

\authorrunning{Xu et al.}

% First names are abbreviated in the running head.
% If there are more than two authors, 'et al.' is used.
%

\institute{
    \hbox{Dept. of Biomedical Engineering, Johns Hopkins University, Baltimore, MD, USA}
    \email{\{txu48,ywu289\}@jh.edu}
    \and
    Johns Hopkins Malaria Research Institute, W. Harry Feinstone Department of Molecular Microbiology and Immunology, Bloomberg School of Public Health, Baltimore, MD, USA
    \email{atripat2@jhu.edu}
    \and
    Dept. of Medicine, Johns Hopkins University School of Medicine, Baltimore, MD, USA
    \email{mippolito@jhu.edu}
    \and
    Dept. of Electrical and Systems Engineering, University of Pennsylvania, Philadelphia, PA, USA
    \email{haeffele@upenn.edu}
}

%==========
%
\maketitle              % typeset the header of the contribution
\begin{abstract}
Deep learning advances have revolutionized automated digital pathology analysis. However, differences in staining protocols and imaging conditions can introduce significant color variability. In deep learning, such color inconsistency often reduces performance when deploying models on data acquired under different conditions from the training data, a challenge known as domain shift. Many existing methods attempt to address this problem via color normalization but suffer from several notable drawbacks such as introducing artifacts or requiring careful choice of a template image for stain mapping. To address these limitations, we propose a trainable color normalization model that can be integrated with any backbone network for downstream tasks such as object detection and classification. Based on the physics of the imaging process per the Beer-Lambert law, our model architecture is derived via algorithmic unrolling of a nonnegative matrix factorization (NMF) model to extract stain-invariant structural information from the original pathology images, which serves as input for further processing. Experimentally, we evaluate the method on publicly available pathology datasets and an internally curated collection of malaria blood smears for cross-domain object detection and classification, where our method outperforms many state-of-the-art stain normalization methods. Our code is available at \href{https://github.com/xutianyue/BeerLaNet}{https://github.com/xutianyue/BeerLaNet}.

\end{abstract}
\section{Introduction}
Pathological examination plays a fundamental role in disease diagnosis, providing critical insights into tissue morphology, cellular abnormalities, and disease progression. However, manual assessment of digital pathology images is labor-intensive, time-consuming, and subject to inter- and intra-observer variability \cite{yeo1993autofocusing,veloso2007interobserver}. Deep learning-driven analysis offers a promising alternative but struggles with a major challenge for model generalizability: color inconsistency in digital pathology. Recorded images can have significant differences in stain appearance even with the same staining protocol, caused by (i) differences in chemical reactions and exposure times of dyes, (ii) variability in sample preparation, and (iii) variable imaging conditions across different scanning hardware \cite{salvi2020stain}. While experienced pathologists can interpret slides despite stain variability, deep learning models are prone to performance degradation \cite{ciompi2017importance}, leading to poor generalization across different staining protocols and acquisition settings. This challenge is closely related to domain adaptation, a key discipline in deep learning that aims to improve generalization across different data distributions. To address this challenge, a strategy called stain color normalization is widely used in domain adaptation methods specific to medical pathology. 

Traditional stain color normalization approaches often attempt to match statistics of the color distribution of test images to those of images in the training domain \cite{reinhard}. Yet, the effectiveness of these methods strongly relies on selecting appropriate representative templates from the training domain to match, which can require prior knowledge of the domain. Moreover, these methods can be generic color transformations that do not account for the underlying physics of histological staining. Other methods leverage algorithms to decompose the color space of a reference image into different dye staining components \cite{macenko,vahadane}. While these methods account for the underlying physics of histological staining and image acquisition, they still rely on a predefined template which the method attempts to match and, in some cases, require additional prior knowledge, such as the absorption spectrum matrix of specific dyes or the number of distinct color components within the images  \cite{vahadane}. More recently, deep learning-based approaches such as generative adversarial networks (GANs) \cite{shaban2019staingan} have been explored for stain style transfer. However, despite their success in aligning color distributions and overall appearance, it has been noted that they can often introduce synthetic artifacts or `hallucinate’ cellular structures, posing risks in medical diagnosis \cite{rahman20213c}.  Other deep learning methods have also proposed to use attention mechanisms to identify relevant color transformations from a variety of standard normalized color spaces, but these methods are also largely generic and do not necessarily account for the relevant image formation process \cite{ke2025learnable}.

To overcome these limitations, we propose the Beer-Lambert Net (\name). 
Our key contributions include: (1) \textbf{Adaptive Stain Disentanglement}: Unlike previous methods, which focus largely on hematoxylin and eosin-stained images, \name extends to arbitrary staining protocols, learning stain-invariant representations of images without requiring any prior knowledge of the staining protocol. (2) \textbf{Trainable and Physics-Informed}: Built on nonnegative matrix factorization (NMF) and algorithmic unrolling, \name enables a data-driven, end-to-end stain decomposition process based on the imaging physics.
(3) \textbf{Flexible Integration}: Designed as a plug-and-play module, \name can be combined with arbitrary backbone networks for downstream tasks, such as object detection or image classification.

\section{Technical Background}
Each stain can be considered a unique element that interacts with light in a specific way. In an RGB image, every stain has distinct absorption characteristics across the three spectral channels, influencing the observed color intensity. %Understanding this interaction is crucial for the next step of stain normalization modeling. 
To mathematically describe how stains modify light intensity in histological images, we rely on the Beer-Lambert law \cite{ruifrok2001quantification}, which models how incident light is attenuated by stains based on their concentration and absorption properties. Specifically, the absorption of stains by histological structures attenuates the incident light on stained biological samples in a specific color spectrum. Given an image $\bar{\X} = [\bar{\x}_0,\bar{\x}_1,...,\bar{\x}_p]\in \mathbb{R}^{c\times p}$, where $\bar{\x}_i \in \mathbb{R}^{c}$ contains the $c$ color intensities for the $i^\text{th}$ pixel ($c$ is typically 3 for RGB images) and $p$ is the number of pixels in the image, then when light $\bar{\x}_0 \in \mathbb{R}^{c} $ illuminates the image, the color intensity at each pixel is given by the Beer-Lambert law;
\begin{equation}
\label{eq:bl_law}
\bar{\mathbf{X}} = (\bar{\x}_0 \mathbf{1}^\top) \odot e^{-\mathbf{SD}^\top}
\end{equation}
where $\mathbf{S} = [\s_1, \s_2,..., \s_r] \in \mathbb{R}^{c \times r} $ is the color appearance matrix whose $i^\text{th}$ column contains the color spectra of the $i^\text{th}$ colored component (e.g., the color of a stain or a naturally colored material in the specimen, such as hemoglobin in red blood cells), $r$ is the number of colored components, $\mathbf{D} = [\d_1,\d_2,...,\d_r] \in \mathbb{R}^{p\times r}$ is the optical density matrix whose $i^\text{th}$ column contains the optical density of the $i^\text{th}$ colored component at each pixel, $\mathbf{1}$ denotes a vector of all ones, and $\odot$ denotes the element-wise matrix product. 

\myparagraph{Matrix factorization for stain separation.}
Based on the Beer-Lambert law, prior approaches have proposed models for stain normalization based on matrix factorization models.  In particular, taking an element-wise logarithm on both sides of \eqref{eq:bl_law} gives
\begin{equation}
    \ln (\bar{\mathbf{X}}) = \ln (\bar{\x}_0) \1^\top-\mathbf{SD}^\top \iff  \ln (\bar{\x}_0) \1^\top - \ln (\bar{\mathbf{X}})  = \mathbf{SD}^\top
\end{equation}
so by estimating a low-rank approximation $\S \D^\top$ (recall the number of columns $r$ in $(\S,\D)$ corresponds to the number of stains/colored materials in the specimen) one could hope to have a reliable decomposition of the image into color and spatial optical density components. For example, the authors of \cite{macenko} propose a model which estimates the image background $\x_0$ via a heuristic and the estimates $\S$ and $\D$ via principal component analysis (PCA). The estimates of $\S$ and $\D$ are then normalized via further heuristics before reforming a normalized version of the original image using the normalized $(\x_0,\S,\D)$ parameters in \eqref{eq:bl_law}.  

While the above model has the benefit of being motivated by the image formation physics, the use of PCA for estimating $\S$ and $\D$ has clear deficiencies, as it requires the columns of the matrices to be orthogonal, which is clearly not realistic for the problem. As a result, the authors of \cite{vahadane} propose a sparse NMF model of the form:
\begin{equation}
    \label{eq:sparse_nmf}
    \min_{\x_0,\S,\D} \frac{1}{2}|| \x_0 \mathbf{1}^\top - \X - \mathbf{SD}^\top||_F^2 + \lambda \sum_{i=1}^r \| \d_i \|_2  \ \ \text{s.t.} \ \ \|\s_i\|_2 = 1, \forall i, \ \ \mathbf{S,D} \geq \mathbf{0}.
\end{equation}
The model in \eqref{eq:sparse_nmf} is motivated by the fact that the color spectra and optical density matrices are known to be nonnegative and additionally adds sparse regularization on the columns of $\D$ via the $\ell_1$ norm based on the common assumption that the relevant histological features of interest are sparse in space.  After estimating the parameters of the model in \eqref{eq:sparse_nmf} the authors then produce a stain-normalized image by applying heuristics to normalize the columns of $\D$. They then use \eqref{eq:bl_law} with the normalized estimate of $\D$ but discard the estimated $\S$ matrix and instead use a fixed color spectra matrix which has been estimated from a template image whose color the method hopes to match.  However, this presents challenges in practice as one must choose a suitable template image and corresponding color spectra matrix and the columns of the estimated $\S$ and $\D$ matrices could be permuted from the corresponding color spectra matrix, resulting in significant color distortions.  Further, the method also requires one to carefully select both the sparse regularization strength, $\lambda$, and the rank of the matrix factorization $r$.  In the following section we present our proposed \name method and describe how it alleviates many of these problems.

\begin{figure}[t]
\center
\includegraphics[width=1\textwidth]{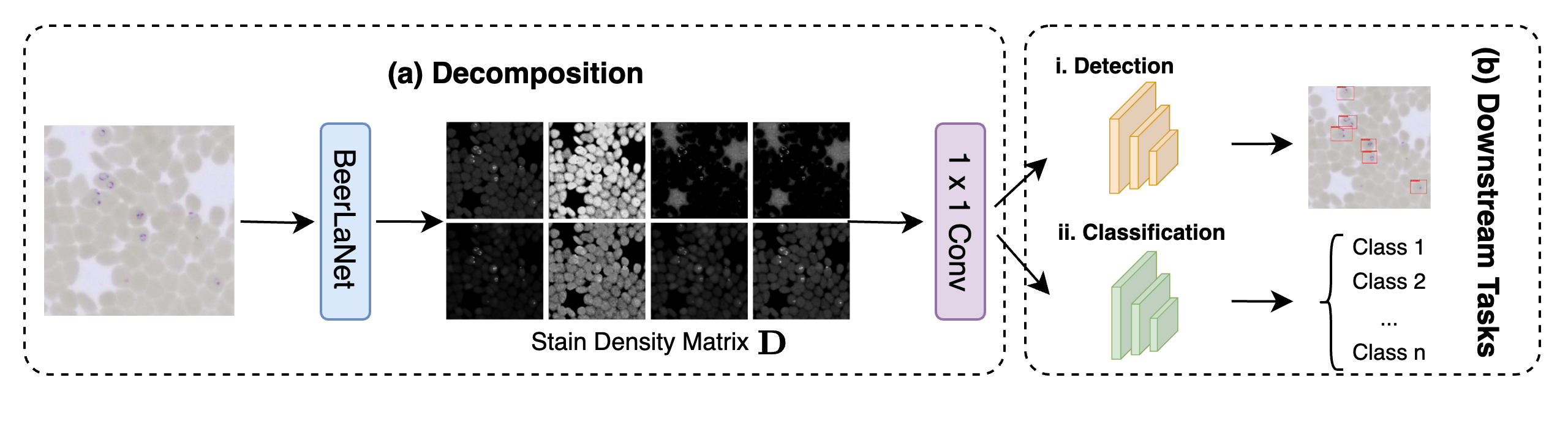}
\caption{Overview of our proposed \name method.}
\label{fig:overview}
\end{figure}

\section{Method}
To derive our method, we begin with similar assumptions to the sparse NMF model described above, but we make several key modifications to mitigate the issues described above and to allow for our method to be incorporated in an end-to-end fashion with any desired deep network model for further downstream tasks.  Specifically, we note that the NMF model in \eqref{eq:sparse_nmf} is non-convex and requires one to solve a non-convex optimization problem due to the $\S \D^\top$ matrix product.  Moreover, the number of columns in $\S$ and $\D$ needs to be specified \textit{a priori}, which can be challenging in applications where the number of colored components is unknown such as when the specimen itself contains colored components in addition to the applied stain (e.g., hemoglobin in red blood cells).  

We address these issues by first leveraging literature on optimizing structured matrix factorization models \cite{haeffele2014structured,haeffele2019structured} and modify \eqref{eq:sparse_nmf} into the following formulation:
\begin{equation}
    \label{eq:main_obj}
    \min_{\mathbf{\x_0,\S,\D}}\frac{1}{2}||\x_0\1^\top-\mathbf{X}-\mathbf{SD}^\top||_F^2 + \lambda \sum_{i=1}^{r}\|\s_i\|_2(\gamma\|\d_i\|_1+\|\d_i\|_2) \ \ \text{s.t}. \ \ \mathbf{S,D} \geq \mathbf{0}.
\end{equation}
Compared to the model in \eqref{eq:sparse_nmf} the key difference is that we have incorporated an additional $\ell_2$ regularization on the columns of $\S$ and $\D$ which is known to promote low-rank solutions due to connections with the variational form of the nuclear norm \cite{haeffele2014structured,haeffele2019structured}.  This allows us to initialize the number of colored components $(r)$ larger than the expected number of components and adapt the rank of the solution to the data.  We then derive our method from an algorithmic unrolling of \eqref{eq:main_obj} which can be integrated in an end-to-end manner with any deep network architecture for further downstream tasks as we describe below.

\myparagraph{Unrolled Network Architecture.} To derive our proposed method, we design an architecture via an algorithmic unrolling of \eqref{eq:main_obj} via alternating proximal gradient descent \cite{combettes2011proximal}.  Specifically, we make the regularization parameters $(\gamma,\lambda)$ and the initialization for $\S$ ($\S_\text{init}$) learnable parameters (with $\x_0$ and $\D$ initialized as all zeros), then our algorithm first updates $\x_0$ in closed form, followed by updates to $\D$ and $\S$ by taking a gradient descent step on the Frobenius norm term in \eqref{eq:main_obj}, followed by solving the proximal operator of the regularization term and nonnegative constraints.  This is then repeated for $K$ unrolled layers, after which we reshape the recovered $\D$ matrix into an $r$-channel image passed through a $1 \times 1$ convolution layer to map the image back to a 3-channel image to allow for passing the image to a backbone network (e.g., YOLO, ResNet) for downstream tasks, with the learnable parameters ($\S_\text{init}, \gamma, \lambda)$ being updated via backpropagation via supervision on the downstream task and with non-negative constraints.  An overview of our approach is depicted in Fig.~\ref{fig:overview}, and the full details of our method are given in Algorithm~\ref{alg:main_alg}.  Note that $(\cdot)_+$ denotes the element-wise ReLU operation and the proximal operator of the regularization/constraints on $\S$ and $\D$ can be shown to be computable by sequentially computing the proximal operator of the non-negativity constraints and $\ell_1$ norm (for the update for $\D$) followed by the proximal operator of the $\ell_2$ norm \cite{haeffele2014structured,haeffele2019structured}.

\begin{algorithm}
\caption{Beer-Lambert Net (\name)}
\begin{algorithmic}
\label{alg:main_alg}
\STATE \textbf{Input:} Image $\mathbf{X} \in \Re^{c \times p}$. \# of unrolled iterations $K$. \# of color components $r$.
\STATE \textbf{Learnable Parameters:} $\gamma \in \Re$, $\lambda \in \Re$, $\S_\text{init} \in \Re^{c \times r}$.
\STATE \textbf{Outputs:} The optical density map matrix $\mathbf{D} \in \Re^{p \times r}$.
\STATE \textbf{Initialize:} $\S = \S_\text{init}$, $\D = \mathbf{0}$, $\x_0 = \mathbf{0}$.
\FOR{$k=1,...,K$}
    \STATE $\x_0 = \frac{1}{p}(\mathbf{X}+\mathbf{SD}^\top)\mathbf{1}$ \ \ \% Closed form update for $\x_0$.
    \STATE $\tau = \frac{1}{\|\mathbf{S}\|_F^2}$ \ \ \% Compute step size for $\D$.
    \STATE $\mathbf{D} = \mathbf{D} - \tau[\mathbf{D S^\top S} + \mathbf{X^\top S} - \mathbf{1} x_0^\top \mathbf{S} ]$ \ \ \% Gradient step of $\|\cdot\|_F^2$ term w.r.t. $\D$.    
    \FOR{$i=1,...,r$} 
        \STATE $\d_i = (\d_i-\lambda \gamma \tau \|\s_i\|_2)_+$ \ \ \% Proximal operator of $\ell_1$ + non-negative constraints.
        \STATE $\d_i = \d_i\left(1-\frac{\min\{\|\d_i\|_2,\lambda \tau \|\s_i\|_2\}}{\|\d_i\|_2}\right)$ \ \ \% Proximal operator of $\ell_2$.
    \ENDFOR
    \STATE $\tau = \frac{1}{\|\mathbf{D}\|_F^2}$ \ \ \% Compute step size for $\S$.
    \STATE $\mathbf{S} = \mathbf{S}-\tau[\mathbf{SD^\top D}+\mathbf{XD}-\x_0 \1^\top \mathbf{D}]$ \ \ \% Gradient step of $\|\cdot\|_F^2$ term w.r.t. $\S$.
    \FOR{$i=0,1,...,r$}
        \STATE $\s_i=(\s_i)_+\left(1-\frac{\min\{\|(\s_i)_+\|_2,\lambda \tau(\gamma \|\d_i\|_1+\|\d_i\|_2)\}}{\|(\s_i)_+\|_2}\right)$ \ \ \% Proximal operator for $\S$.
    \ENDFOR
\ENDFOR
\end{algorithmic}
\end{algorithm}

\section{Experimental Performance}

\begin{figure}[t]
    \centering
    \includegraphics[width=\textwidth]{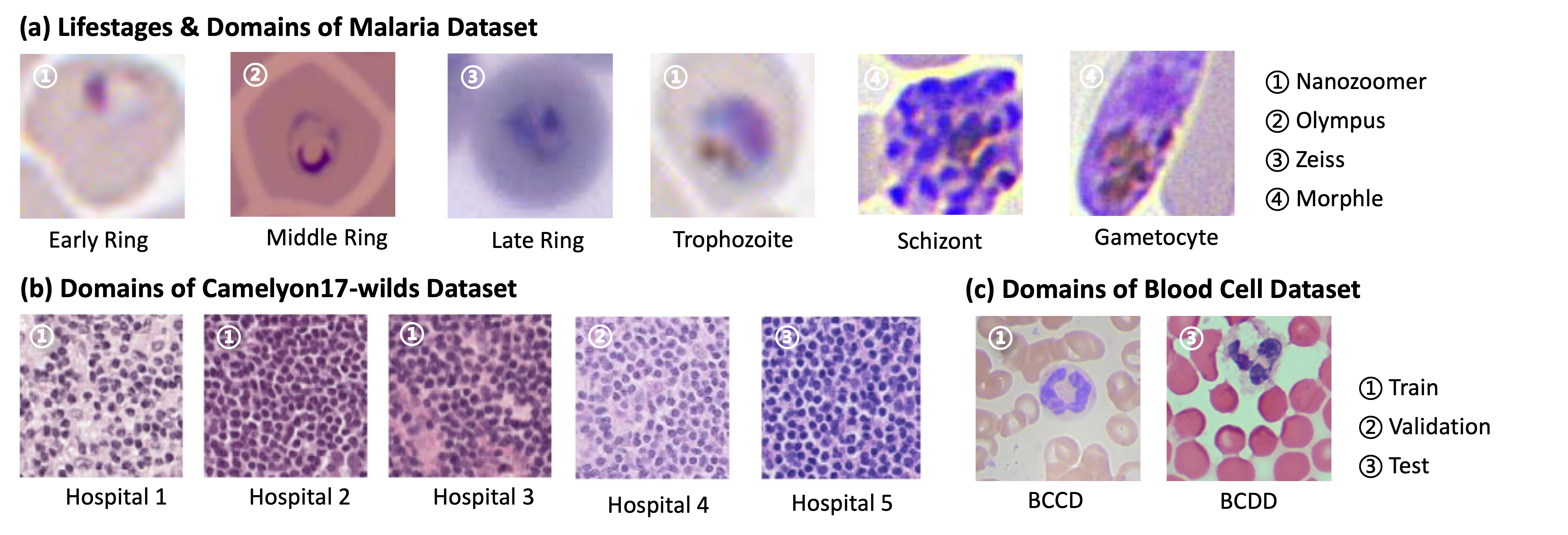}
    \caption{Example images from our tested datasets.} \label{dataset_visulization}
\end{figure}

We evaluate our method on a variety of modalities for two diagnostic tasks: object detection and image classification. Here, we detail the modalities and their respective role in our study, and example images showing the variance in domain between training and testing data are displayed in Fig~\ref{dataset_visulization}.

\myparagraph{Malaria Parasite Detection and Classification.} We used images of malaria blood smears for both detection and classification tasks. For detection, we used a public malaria parasite detection dataset by the authors of \cite{guemas2024automatic}, which comprises 24,720 May Grunwald-Giemsa (MGG)-stained thin blood smear images with annotations across four categories: white blood cells, red blood cells, platelets, and parasites (including \textit{Trypanosoma brucei}, and erythryocytes infected by \textit{Plasmodium falciparum}, \textit{P. ovale}, \textit{P. vivax}, \textit{P. malariase}, and \textit{Babesia divergens}). We additionally collected and manually labeled a test dataset comprising 264 thin blood smear images captured from a Zeiss Axioscan microscope.

For classification, we assembled a diverse dataset from four microscopy platforms, encompassing both manual and high-throughput imaging systems with varying numerical apertures (Hamamatsu NanoZoomer, Zeiss Axioscan, Olympus CX43, Morphle Hemolens). The source materials included Giemsa-stained thin blood smears obtained from multiple origins: a clinical trial, laboratory-prepared slides with spiked parasites, and de-identified slides from a malariology course collection. Classification labels were independently assigned by two authors (AKT, MMI), with any disagreements resolved through joint review by both. The dataset consists of single-cell cropped images infected by \textit{P. falciparum}, with a training set with 2,486 samples, and two test sets with 343 and 261 samples, each from a distinct domain (imaging platform). The classification task was to label detected parasites by lifestage: early ring, middle ring, late ring, trophozoite, schizont, and gametocyte, as shown in Fig.~\ref{dataset_visulization}(a).

\myparagraph{Breast Cancer Classification.}
The Camelyon17-WILDS dataset \cite{bandi2018detection} is a large-scale histopathology dataset for evaluating domain adaptation in medical image analysis. It consists of $96\times 96$ image patches extracted from lymph node whole slide images, with tumor presence labeled in the central $32\times32$ region. The dataset spans five hospitals, introducing domain shifts in staining and imaging conditions. It contains tumor/non-tumor classifications, with 302,436 training, 34,904 validation, and 85,054 test samples.  Here we train on the 3 training datasets and perform inference on the testing and validation datasets.

\myparagraph{Whole Blood Cell Detection.}
We used two additional public datasets for detecting various blood cell types from whole blood smears: the Blood Cell Count and Detection (BCCD) dataset \cite{Shenggan} containing 366 blood smear images for training, and the Blood Cell Detection Dataset (BCDD) \cite{bcdd} with 100 blood smear images for testing. 

\myparagraph{Implementation Details.} For \name, we set the number of colored component $r$ to 8 and internal iterations to 10. For detection tasks, we implemented our method with a YOLOv8 backbone \cite{Jocher_Ultralytics_YOLO_2023}. The training parameter batch size was set to 8, and the number of epochs was set to 50. The optimizer was automatically chosen by YOLO with an initial learning rate of 0.01. For classification, we trained the model using Adam optimization with a batch size of 128 and a ResNet-18 backbone \cite{he2016deep}. The \textit{P. falciparum} malaria parasite dataset was trained for 20 epochs and Camelyon-17 WILDS for 50 epochs, both with a learning rate of 1e-4. For the malaria classification dataset, we also applied a two-step denoising pipeline due to compression artifacts from the imaging platform: a median filter with kernel size of 11, followed by a Gaussian blur with kernel size of 21 and $\sigma=1$. All experimental results are reported as averages across 3 random seeds. % to ensure reproducibility. \footnote{The full implementation code will be made publicly once paper is accepted.}

\myparagraph{Comparison Methods \& Evaluation Metrics.} We compared our proposed method with 3 classical histology normalization methods—Reinhard \cite{reinhard}, Macenko \cite{macenko}, and Vanhadane \cite{vahadane}—and two deep learning-based methods—Stain-GAN \cite{shaban2019staingan} and LStainNorm \cite{ke2025learnable}. For those methods requiring a template, we randomly select one image from the training dataset as the template for normalization. The inference results are evaluated by mAP$_{50}$ and mAP$_{50-95}$ on the detection task.  For classification we report classification accuracy (Acc), and for the malaria parasite classification task we also report a relaxed accuracy (RAcc) which considers a classification successful if the manual and predicted labels are within one lifestage of each other for the early/middle/late ring stages (e.g., if the manual label is `early ring', then predictions of `early ring' or `middle ring' are considered correct) to reflect the fact that the parasite growth is a continuous process and the discretization into 3 stages is somewhat arbitrary. In addition, to compute an average performance across all tasks and metrics, we also introduce the average percent underperformance (APU)  which computes the percent difference between the current method and the best-performing method for a particular task+metric (percent underperformance) and then averages this across all task+metric combinations for either detection or classification. For example, for each entry in Table~\ref{tab2}, we compute the percent difference with the maximum value in the column and then average the percent differences across the 4 entries in the row for detection and classification.

%% Original Table: Detection and classification merged in width
\begin{table}[t]
\centering
\fontsize{8pt}{10pt}\selectfont
\caption{Comparison of Stain Normalization Techniques. %APU: Average percent underperformance on all datasets. 
The best and the second-best results are \textbf{boldfaced} or starred (*), respectively. (C17 denotes Camelyon17-WILDS)}
\label{tab2}
\begin{tabular}{l|c c c c c|c c c c c}
\hline
Task & \multicolumn{5}{c|}{Detection (YOLOv8)} & \multicolumn{5}{c}{Classification (ResNet18)} \\ \hline
Dataset & \multicolumn{2}{c}{Malaria} & \multicolumn{2}{c}{Whole Blood Cells} & & \multicolumn{2}{c}{Malaria} & C17$_\text{test}$ & C17$_\text{val}$ & \\ 
 % & & & & & & \multicolumn{1}{c|}{Test} & \multicolumn{1}{c}{Val}\\ \hline
Metrics (\%) & mAP$_{50}$& mAP$_{50-95}$ & mAP$_{50}$ & mAP$_{50-95}$ & \cellcolor{gray!20}APU & Acc & RAcc &Acc& Acc & \cellcolor{gray!20}APU \\ \hline
Baseline & 91.03 & 52.00 &65.20 & 36.70 & \cellcolor{gray!20}18.10& 21.32 & 45.59 & 85.21 & 83.38 & \cellcolor{gray!20}31.70\\  \cdashline{1-11}

Reinhard &90.17 & 51.47 &79.53 & 44.03 & \cellcolor{gray!20}11.17 & 29.34 & 62.30 & 94.55 & 91.36* & \cellcolor{gray!20}18.36 \\ 
Macenko &72.03 & 39.27 &65.43 & 35.27 & \cellcolor{gray!20}29.27 & 29.59 & 73.54 &\textbf{95.92}&85.77&\cellcolor{gray!20}16.27 \\ 
Vahadane & 92.10*& 55.87* & 57.57  & 31.30 & \cellcolor{gray!20}20.76 & 38.81* & 81.04* & 95.85*& 86.43 & \cellcolor{gray!20} 9.31* \\  \cdashline{1-11}

StainGAN &81.67 & 37.70 & \textbf{89.60} & \textbf{53.97} & \cellcolor{gray!20}12.02& 31.46 & 70.94 & 94.28 & 90.77 &\cellcolor{gray!20} 15.12 \\ 
LStainNorm &91.80 & 53.67 &85.83 & 50.00 & \cellcolor{gray!20}5.25*& 21.17 & 61.82 & 93.23& \textbf{92.56} &\cellcolor{gray!20} 22.72 \\ \hline
\textbf{\name} &\textbf{95.07} & \textbf{57.10} &86.80* & 51.33* & $ \textbf{\cellcolor{gray!20}2.00}$ & \textbf{48.66} & \textbf{90.33} & 91.36 & 90.09 &\textbf{\cellcolor{gray!20}1.86}\\ \hline
\end{tabular}
\end{table}

\myparagraph{Results.} Our experiment results are presented in Table~\ref{tab2}.  Here we note that \name achieves the best performance for both the malaria parasite detection and classification tasks and the second highest performance on the whole blood cell detection task, while also having excellent consistency across all tasks and datasets.  For example, on the Camelyon-17 WILDS classification task, the Macenko method achieves the best performance on one dataset (C17$_\text{test}$) but drops 10 percentage points on the other Camelyon-17 WILDS dataset (C17$_\text{val}$).  Likewise, for virtually all comparison methods, a large drop in performance can be observed for one of the tasks/datasets.  In contrast, even when \name is not the top method for a particular dataset, it still performs very competitively, which is reflected in the APU metric, where \name significantly outperforms the comparison methods and quantifies the consistently strong performance of \name.  We conjecture this discrepancy in performance across tasks/datasets observed in comparison methods can potentially be attributed to greater domain variations in the malaria dataset compared to Camelyon17-WILDS and whole blood cell datasets (see Fig.~\ref{dataset_visulization}), datasets with relatively minor color shifts compared to those observed in the malaria dataset.  Methods with a generic design that do not incorporate stain-specific characteristics may perform well on small domain shift data (Camelyon17-WILDS, whole blood cells) but fail to generalize to the large color shifts observed in the malaria dataset.  This highlights the flexible nature of \name to perform robustly in a variety of conditions without prior knowledge of imaging or staining protocols.

\section{Discussion and Conclusion}

Here we have presented \name, an adaptive stain normalization method that enhances cross-domain generalization in digital pathology through physics-based principles and algorithmic unrolling. Our experimental results demonstrate that \name consistently outperforms traditional normalization techniques across multiple tasks and datasets. The key advantages of \name include end-to-end trainability, template-free operation, physics-informed decomposition, and flexibility with various staining protocols. Potential future work could focus on exploring additional tasks such as segmentation and other histopathological domains. 

% =========================================
% =========================================
\begin{credits}

\subsubsection{\ackname} This work was supported by NIH NIAID grant R21AI16936, Bloomberg Philanthropies through the Johns Hopkins Malaria Research Institute, and the Johns Hopkins Institute for Data-Intensive Engineering and Science.

\subsubsection{\discintname}
The authors have no competing interests to declare that are
relevant to the content of this article.
\end{credits}
%
% ---- Bibliography ----

\bibliographystyle{splncs04}
\bibliography{Paper-4658}

\end{document}